\documentclass{article}
\usepackage[utf8]{inputenc}
\usepackage{amsmath}
\usepackage{graphicx}
\usepackage{amssymb}
\usepackage{authblk}
\usepackage{hyperref}

\title{Calibrating Ensembles for Scalable Uncertainty Quantification in Deep Learning-based Medical Segmentation}
\author[1,2]{Thomas Buddenkotte}
\author[2,3]{Lorena Escudero Sanchez}
\author[3,4,5]{Mireia Crispin-Ortuzar}
\author[2,3,6]{Ramona Woitek}
\author[2,3]{Cathal McCague}
\author[3,4,5]{James D. Brenton}
\author[7]{Ozan {\"O}ktem}
\author[2,3]{Evis Sala}
\author[2,3,8]{Leonardo Rundo}
\affil[1]{Department of Applied Mathematics and Theoretical Physics, Univeristy of Cambridge, United Kingdom}
\affil[2]{Department of Radiology, University of Cambridge, United Kingdom}
\affil[3]{Cancer Research UK Cambridge Centre, University of Cambridge, United Kingdom}
\affil[4]{Cancer Research UK Cambridge Institute, University of Cambridge, United Kingdom}
\affil[5]{Department of Oncology, University of Cambridge, Cambridge, United Kingdom}
\affil[6]{Medical Image Analysis \& Artificial Intelligence (MIAAI), Department of Medicine, Danube Private University, Krems, Austria}
\affil[7]{Department of Mathematics, KTH Royal Institute of Technology, Sweden}
\affil[8]{Department of Information and Electrical Engineering and Applied Mathematics, University of Salerno, Fisciano (SA), Italy}
\date{}                     %% if you don't need date to appear
\begin{document}

\maketitle
\begin{abstract}
    Uncertainty quantification in automated image analysis is highly desired in many applications. Typically, machine learning models in classification or segmentation are only developed to provide binary answers; however, quantifying the uncertainty of the models can play a critical role for example in active learning or machine human interaction. Uncertainty quantification is especially difficult when using deep learning-based models, which are the state-of-the-art in many imaging applications. The current uncertainty quantification approaches do not scale well in high-dimensional real-world problems. Scalable solutions often rely on classical techniques, such as dropout, during inference or training ensembles of identical models with different random seeds to obtain a posterior distribution. 
    In this paper, we show that these approaches fail to approximate the classification probability. On the contrary, we propose a scalable and intuitive framework to calibrate ensembles of deep learning models to produce uncertainty quantification measurements that approximate the classification probability. On unseen test data, we demonstrate improved calibration, sensitivity (in two out of three cases) and precision when being compared with the standard approaches. We further motivate the usage of our method in active learning, creating pseudo-labels to learn from unlabeled images and human-machine collaboration. 
\end{abstract}

\section{Introduction}

Uncertainty quantification (UQ) in image classification and segmentation is an actively studied area of research \cite{UQ_review}. Many applications across various imaging tasks can benefit from having access to the certainty of a machine learning model instead of just considering the binary answer of the machine learning system. For example, the work in \cite{human_ai} used a machine learning system that classified cities from images of famous sights and calibrated it for improved human machine interaction. A second important application of UQ in machine learning is active learning \cite{AL_survey}. The ultimate goal is efficient use of manual labelling resources \cite{total_segmentator}. Active learning aims at finding images from a pool of unlabeled images that give the largest performance increase when being labelled and added to the training data. Typically, model uncertainty and sample diversity are the main factors considered in such selection processes. UQ can also be applied to directly learn from unlabeled images as suggested in \cite{UQ_semisupervised1, UQ_semisupervised2}. 

Since the invention of the U-Net architecture \cite{unet} and the nnU-Net framework \cite{nnUNet}, medical segmentation problems have been dominated by deep learning-based algorithms \cite{Cuttingedge3M}. Currently, deep learning models based on the U-Net architecture or the nnU-Net framework are the state-of-the-art across a large amount of medical segmentation tasks. However, the labelling of large 3D scans can take up to hours of expert's time per scan. Consequently, the corresponding datasets typically only have a few hundred samples instead of thousands or millions as in other imaging tasks. Naturally it is desirable to integrate active learning \cite{AL_survey, total_segmentator}, human-machine collaboration \cite{human_ai, human_AI_collaboration} or learning from unlabeled images \cite{UQ_semisupervised1, UQ_semisupervised2, pseudo_labels, pseudo_labels2, medical_self_supervised, self_supervised_wild} into the development cycle of deep learning-based segmentation systems and large annotated datasets.

The main challenge in applying UQ to deep learning is that deep neural networks tend to produce overconfident predictions. It is tempting to interpret the output of the softmax layer as a probability, however it has been shown that these are overconfident and do not represent the certainty of the model well \cite{overconfident}. This is especially prevalent in medical segmentation tasks as here state-of-the-art models often use a combination of entropy-based and overlap-based loss functions \cite{Cuttingedge3M}, which force the networks to produce almost binary predictions. Many alternatives---such as variational inference, Bayesian backpropagation, or Bayesian neural networks---have been suggested for the use in UQ tasks \cite{UQ_review}, however most of these methods are difficult to apply in large scale 3D networks \cite{scalable_UQ}. The two most popular approaches that can be employed in these problems are: (i) to use Dropout both during training and inference, and (ii) to train ensembles of identical models with different random seeds \cite{scalable_UQ}.

In this paper, we propose a scalable and intuitive framework for UQ in 3D medical image segmentation. The framework simply suggests training models ranging from highly sensitive to highly precise and combine their outputs with appropriate weightings. We demonstrate that our framework is superior in estimating classification probabilities when being compared to well-established methods, such as inference Dropout and the traditional ensembling. We further motivate its adoption in active learning \cite{AL_survey, total_segmentator} and creating pseudo-labels to learn from unlabeled images \cite{pseudo_labels, pseudo_labels2} and human-machine collaboration \cite{human_ai, human_AI_collaboration}.

This manuscript is organized as follows. Section 2 motivates and introduces our suggested approach. Section 3 describes the datasets and deep learning models used throughout the manuscript. Section 4 shows results on comparing our method to two established approaches. Section 5 motivates and describes potential ways of applying this method in a clinical setting or in creating improved deep learning algorithms. Section 6 summarizes the paper and discusses potential next steps.

\section{Calibration of Ensembles for Uncertainty Quantification}\label{sec:calibrating_ensembles}

Let us consider a one-dimensional classification task as a motivational example. Consider an equal number of samples from the two distributions $p_0=\mathcal{N}(-1,1)$ and $p_1=\mathcal{N}(1,1)$ are given and the task to determine whether some sample $x$ came from the distribution $p_0$ or $p_1$. Naturally one would suggest the function:
\[ c(x)=
\left\{
\begin{array}{ll}
0 & x \leq 0 \\
1 & \, \textrm{otherwise} \\
\end{array},
\right.
\]
as a classifier that achieves maximal accuracy. In this problem, it is expected that the uncertainty of the classification is high close to the decision boundary in $x=0$. As in this problem the densities of the two distributions are known, the classification probability can be computed analytically via

\[ \mathbf{P}(x\in C_1) = \frac{p_1(x)}{p_0(x) + p_1(x)}.\]

\begin{figure}
    \centering{
    \includegraphics[scale=0.3]{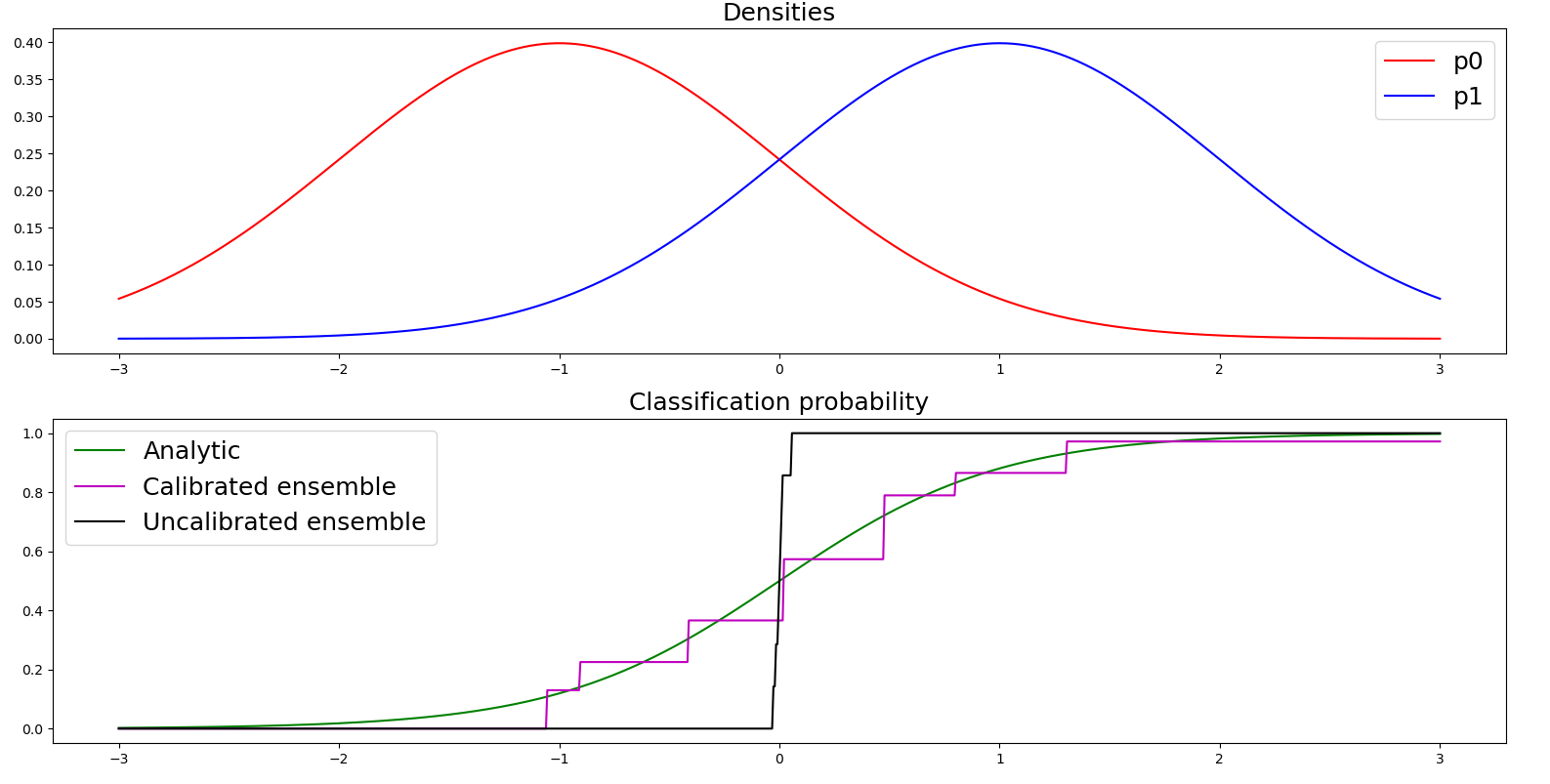}}
    \caption{A minimal example of classification functions and classification probability for a one-dimensional feature space. The example demonstrates why training identical models with different random seeds fails to approximate the classification probability.}
    \label{fig:minimal_example}
\end{figure}

A trained classifier for this problem would not place the decision boundary exactly at $x=0$, but only close by due to the early stopping of the training. As a consequence, ensembles of such models would only express uncertainty close around $x=0$, but would not approximate the classification accuracy well as shown in Figure \ref{fig:minimal_example}. These plots suggest why ensembles of identical models trained with different random seeds are not suited to approximate the classification probability. All decision boundaries are distributed close around 0, where a step function is also required to have jumps further away from 0 to approximate the classification probability well. Thus, it is necessary to train classifiers with shifted decision boundaries. Additionally, it is necessary to find suitable weightings for the different models to adjust the step heights and to finally produce a good approximation of the classification probability.
%To do so it is necessary to shift the decision boundaries of the classifiers and find suitable weightings for those models in the ensemble.

We suggest creating models with a large variety of decision boundaries by altering the importance of sensitivity and precision throughout the ensemble. We achieve this by using parametric loss functions for the calibration of models in the ensemble. For this, we modify the widely used sum of cross entropy and soft dice loss in the following way:

\begin{align}
    L_{w}(y,p) &= L^{ce}_w(y,p) + L^{DSC}_w(y,p)\nonumber\\ 
    L^{ce}_{w}(y,p) &= \frac{1}{N}\sum_{i=1}^N y_{0,i}\log(p_{0,i}) + \sum_{c=1}^C \exp(-w_c) y_{c,i}\log(p_{c,i})\label{eq:weighted_loss}\\ 
    L^{DSC}_{w}(y,p) &= \frac{1}{C}\sum_{c=1}^C 2\frac{TP}{TP + \sigma(w_c)FP + \sigma(-w_c)FN},\nonumber
\end{align}
where $C$ is the number of foreground classes, $y$ is the ground truth given in one hot encoding, $p$ is the softmax output of the neural network, $\sigma$ is the sigmoid function and $TP$, $FP$, and $FN$ are the number of true positives, false positives and false negatives. It is to be noted that the original sum of cross entropy and dice loss can be obtained by choosing $w=0$. In this formulation negative weights promote sensitivity and positive weights promote precision.

To obtain meaningful estimations for the classification probability, one needs to compute a weighted sum of the models trained with differing loss function weights $w$. However, in contrast to the minimal example from Figure \ref{fig:minimal_example}, we cannot compute the classification probability analytically and need to rely on other ways to determine meaningfulness. For this reason, we suggest to compare \textit{the estimated probabilities with the prevalence of foreground voxels}. The rational is that these two quantities should match. For example, if 20\% of the voxels marked with a probability of 0.2 contain foreground, it can be argued that 0.2 was a reasonable estimate for the classification probability. In the calibrated ensemble we suggest achieving this by matching the two quantities on the training set. In a first step we train the model following a cross-validation procedure to obtain one (non-overfitted) prediction for each image in the training set. Next, we setup up a linear system to determine the coefficients in the linear combination. For this, let $s_{w_k}$ be the segmentation models of the ensemble with loss weights $w_k$. Since only binary segmentation models are considered in this approach, the vector $(s_{w_0}(x)_i,\dots,s_{w_n}(x)_i)$ has only binary entries for each image $x$ and voxel $i$. Given the set $B$ of all possible binary vectors and a set of images $X$, the linear system 
\begin{equation}
  \begin{aligned}
    \sum_{x\in X}\sum_i g_i & \mathbb{I}((s_{w_0}(x)_i,\dots,s_{w_n}(x)_i)=b) = \\
    &\sum_k a_kb_k \sum_{x\in X}\sum_i \mathbb{I}((s_{w_0}(x)_i,\dots,s_{w_n}(x)_i)=b) \quad b\in B \label{eq:linear_system}%\nonumber
  \end{aligned}
\end{equation}
can be used to determine the coefficients $a_k$ of the calibrated ensemble, where $\mathbb{I}$ is the indicator function (1 if the expression is true, 0 otherwise).
The heatmap, which approximates the classification probability in each pixel, can finally be computed as
\begin{equation}
    h = \sum_k a_k        
          s_{w_k}.\label{eq:linear_combination}
\end{equation}

\section{Materials and Methods}

\subsection{Datasets}
The experiments were conducted on two independent datasets: (i) a dataset of CT images of high grade serous ovarian cancer patients, and (ii) a public dataset of CT images of kidney tumors\footnote{https://kits21.kits-challenge.org/}\cite{kits_challenge}.

The ovarian cancer data consisted of four datasets in total. All scans were collected retrospectively after receiving the patient's consent for the usage in research. For training we used a dataset of 276 scans from 157 patients that were referred from various clinics to Addenbrooke's hospital in Cambridge, UK, for diagnosis and treatment. From all patients a pre-treatment scan was available and for those treated with neoadjuvant chemotherapy and additional post-treatment scan was available as well. The test data was obtained by merging three datasets. The first one contained 104 scans from 53 patients, which received neo-adjuvant chemotherapy, obtained at St. Bartholomew's (Barts) hospital in London, UK. The second dataset was obtained by the Apollo network in the US and consisted of 51 scans from 51 patients which received immediate primary surgery. The last dataset consisted of 20 scans from 20 patients from the cancer imaging archive (TCIA), which were also acquired in the US from patients who also received immediate primary surgery.

All scans were either manually segmented by a trainee radiologist and subsequently reviewed and corrected by a consultant radiologist, or directly segmented by the consultant radiologist. For the 104 scans obtained at Barts hospital in London, both the segmentations reviewed and corrected by the consultant radiologist (ground truth) and the unreviewed segmentations produced by the trainee radiologist (trainee segmentations) were available. While this disease is regularly expressed in many different sites across the whole abdomen, we focus only on the main two disease sites which are the pelvis/ovaries and the omentum, as they are by far the largest and most frequent sites.

Additionally, we used the Kidney Tumor Segmentation 2021 (KiTS21) dataset\footnote{https://kits21.kits-challenge.org/} \cite{kits_challenge}. The dataset consists of 300 publicly available scans to be used for model training and a hidden test dataset. Both kidneys and all prevalent kidney tumors and cysts were segmented by three radiologists independently. At least one tumor was visible in each scan. We decided to focus only on the kidney tumors for simplification and used the majority vote of the three prevalent masks for training and testing. The publicly available data was randomly split in 60\% training (n=180) and 40\% test scans (n=120).

\subsection{Deep learning model}

The implementation is based on PyTorch 1.9 and was coded from scratch, i.e., not relying on public open libraries. However, our code design and hyper-parameter closely follow the implementation in the nnU-Net framework \cite{nnUNet}. All scans were first interpolated to a uniform voxel spacing of $0.8\times0.8\times5 \text{mm}^3$, which resulted in higher DSC compared to the median voxel spacing, which was suggested by nnU-Net. The resizing was followed by windowing the image by the 99.5 and 0.5 percentile of the gray values of voxels that contain foreground and Z-score normalization. The augmentation was applied as suggested by nnU-Net \cite{nnUNet}. The architecture was chosen to be U-Net with a ResNet \cite{resnet, nf_resnet} decoder using 1, 2, 6, and 3 blocks per stage and 32, 63, 128, and 256 filters per stage. For the ovarian cancer segmentation we used a single multi-class U-Net to segment both disease sites simultaneously. For the kidney data we used a single-class U-Net for only the tumors. Additionally, we only sampled patches centered at randomly chosen voxels inside the kidney and excluded voxels outside the of kidneys by using a mask in the loss function. The training was performed with a standard stochastic gradient optimizer using Nesterov's momentum of factor 0.98, a weight decay of $10^{-4}$ and a linear ascent plus cosine decay learning rate schedule with maximum rate 0.02. We used a batch size of 4 for all experiments where one sample of each batch was forced to be centered at a randomly chosen foreground voxel. Full 3D volumes were predicted using the Sliding Window approach with an overlap of 0.5 patches and a Gaussian weighting as suggested in \cite{nnUNet}. The post-processing was performed by simply applying the argmax function to the softmax outputs of the network.

To reduce the training cost, we applied progressive learning as suggested in \cite{effnetv2}. Here the training is split in four equally long stages. The amount of voxels contained in a sample and the magnitude of the augmentations was reduced by a factor of 4, 3, and 2 in the first, second, and third stage of the training. This alone reduced the training time by roughly 40\%. However, preliminary experiments also showed that full retraining from scratch is not needed when changing the loss weight $w_k$ in the loss function. In this case it is sufficient to restart the training after the third stage of the progressive learning and finish the training with the new choice of $w_k$.

For the calibrated ensemble, we aimed at sampling the loss weights $w$ symmetrically around the parameter $w^{DSC}$ that maximizes the mean DSC. Using cross-validation we determined $w^{DSC}$ (-0.5, -1.5, and -1.0 for pelvic/ovarian disease, omental disease and kidney tumors respectively) and chose the loss weights of the calibrated ensembles to be \mbox{$\{w^{DSC}-k|k=-3,-2,\dots,3\}$}. For each of these loss function weights a full cross-validation was trained to determine the coefficients $a_k$ as described in equation \ref{eq:linear_system}. Finally, for each $w_k$ ensembling over all models in the cross-validation predictions was applied on the test set before computing the heatmap as suggested by equation \ref{eq:linear_combination}.

As a baseline to compare our approach with, we trained an (uncalibrated) ensemble seven models trained with random seeds and a network using Dropout with drop probability $p=0.1$ after each convolution (except for the logits). Both were trained using the full training data instead of relying on cross-validation splits, embracing the weighted loss function from Equation \ref{eq:weighted_loss} with weight $w^{DSC}$. The Dropout model was evaluated seven times on the test data with Dropout in training mode. The heatmaps were computed as the pixel-wise mean over the seven predictions for both the uncalibrated ensembles and the Dropout model.

The implementation is available at \url{https://github.com/ThomasBudd/ovseg}. Please follow the documentation and manual to learn how to use the library for new data and apply the suggested method to your segmenation problem.

\section{Results}

\begin{figure}
    \centering
    \includegraphics[scale=0.28]{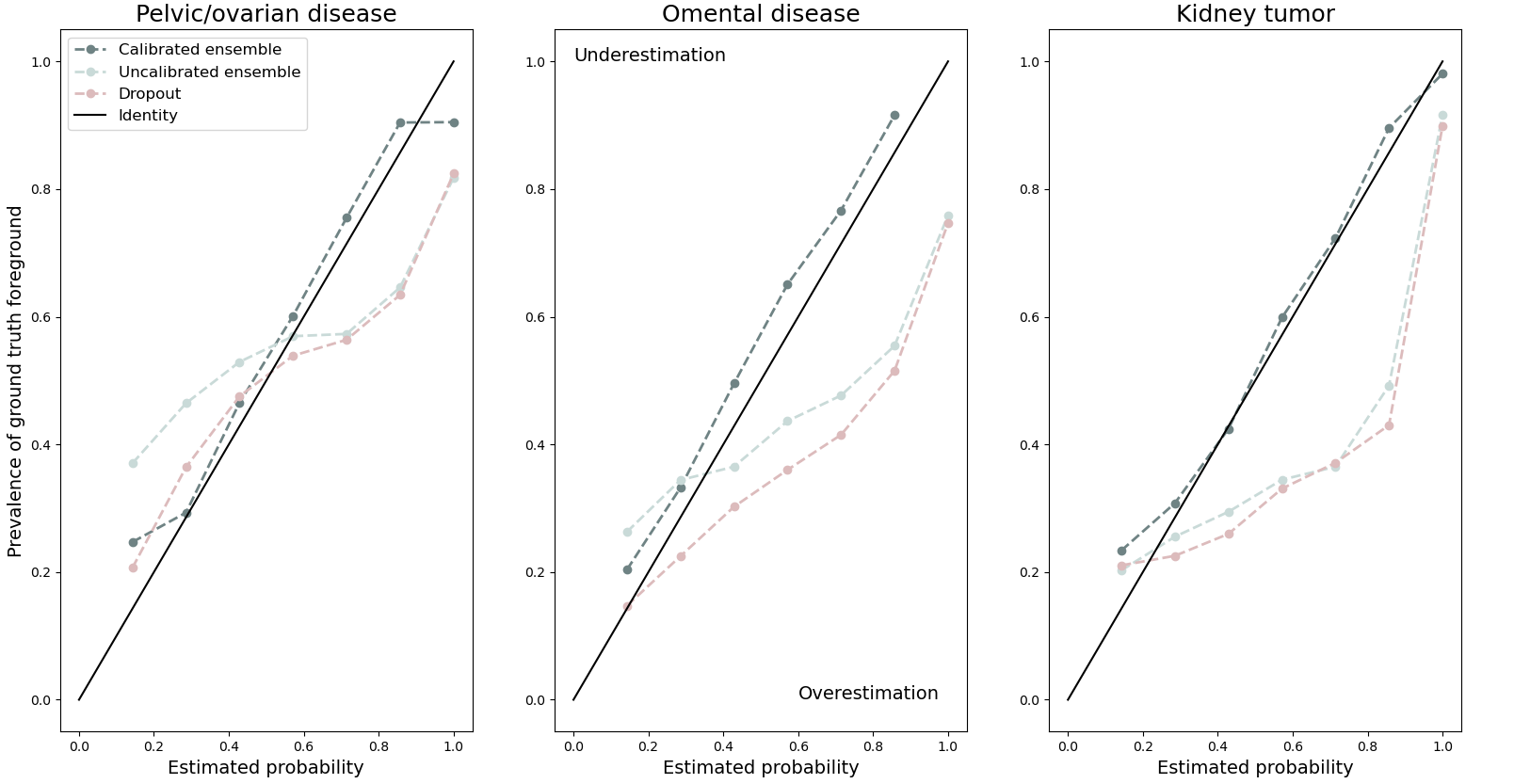}
    \caption{Calibration curves of the different uncertainty quantification approaches (dashed lines) compared with perfect calibration (solid line). The area above and below the solid line correspond to underestimating and overestimating probabilities respectively as marked in the middle plot.
    %Validation of estimated classification probabilities. It can be observed that the probabilities provided by the uncalibrated ensemble and inference dropout network deviate from the identity function, while the calibrated ensemble provides a better approximation.
    }
    \label{fig:p_vs_p}
\end{figure}

To check the meaningfulness of the heatmaps the predicted probabilities were compared with the prevalence of foreground voxels in the test set. As previously described in Section \ref{sec:calibrating_ensembles}, a probability estimation can be considered meaningful if these two quantities match. If the prevalence of foreground voxels is higher/lower than the predicted probability, the uncertainty quantification approach underestimates/overestimates the risk. We used Parzen Windowing \cite{parzen} with a linear kernel to compute the pixel-wise classification probability as the ratio of ground truth voxels per bin. The results are shown in Figure \ref{fig:p_vs_p}. It is clearly visible that the calibrated ensemble has a better calibration on external test for all three segmentation tasks, when being compared to uncalibrated ensembles and dropout networks. The latter two have instead a tendency for overestimation in regions of high probability and underestimation in regions of low probability.

\begin{figure}
    \centering
    \includegraphics[scale=0.5]{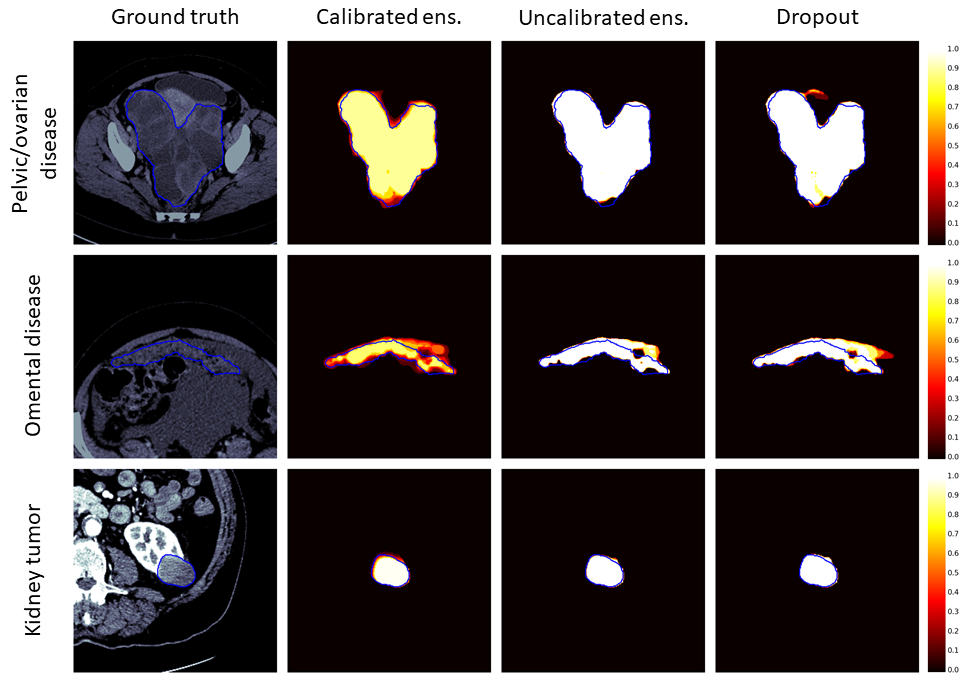}
    \caption{Visual comparison of the heatmaps created by different approaches. The uncalibrated ensemble and Dropout heatmaps are almost binary and falsely include and exclude voxels with high certainty. This is considerably improved in the heatmaps of the calibrated ensemble.}
    \label{fig:compare_heatmap}
\end{figure}

Figure \ref{fig:compare_heatmap} shows heatmaps of the three different approaches. The heatmaps produced by uncalibrated ensembles and Dropout models express uncertainty in relatively few voxels compared to our suggested approach. Instead, some voxels were falsely included or excluded with very high certainty. In contrast to this, the calibrated ensemble captures almost all foreground, even if with low probability in some cases. Voxels outside the ground truth segmentation were typically annotated with a low probability. To quantify this, we computed the sensitivity of the union and the precision of the intersection of all binary segmentation masks for each approach. The results can be found in Table \ref{tab:sens_prec}. Overall, the calibrated ensemble achieves the highest values except for the sensitivity for the omental disease, where the Dropout model achieves a marginally higher score.

\begin{table}[ht!]
\centering
\begin{tabular}{l | c || c | c | c }
                    &        & Uncalibrated & Calibrated & Dropout\\
     Disease        & Metric & ensemble     & ensemble   & model \\ \hline\hline
     Pelvic/ovarian & Sens   & 89.0         & 84.6       & 86.9 \\
     disease        & Prec   & 90.4         & 81.7       & 82.3 \\ \hline
     Omental        & Sens   & 75.5         & 72.4       & 75.6 \\
     disease        & Prec   & 91.6         & 75.9       & 74.7 \\ \hline
     Kidney         & Sens   & 95.5         & 94.3       & 94.9 \\
     tumor          & Prec   & 98.1         & 91.6       & 89.9
\end{tabular}
\caption{Sensitivity of the union and precision of the intersection of all binary segmentation masks produced by the different UQ approaches. The calibrated ensemble achieves overall the highest scores except for the sensitivity of the omental disease.}
\label{tab:sens_prec}
\end{table}

%Examples of manual contours and heatmaps created by our approach can be found in Figure \ref{fig:heatmap_examples}. Comparing the first and second row, it can be observed that greater uncertainty occurs in cases with poor contrast compared to cases with clearly visible disease boundaries. The third row demonstrates the difficulty of segmenting small lesions. The slightly larger connected components were segmented with greater certainty than the very small lesions in the very left and right of the omental cake. The fourth row shows that also kidney tumors could be segmented with high certainty by the approach. Compared to the three manual segmentations, the deep learning model struggles more in finding the edges of the tumor.  

%\begin{figure}
%    \centering
%    \includegraphics[scale=0.6]{heatmap_examples.png}
%    \caption{Visual demonstration of the suggest UQ approach. The images in the first column show solely the reconstruction, where the manual segmentation was added in the second column. The three manual segmentations that were available for the kidney data are visualized as a heatmap. The last two columns show the heatmaps created by the suggested approach.}
%    \label{fig:heatmap_examples}
%\end{figure}

\section{Applications to Active Learning and Processing of Unlabeled Images}
In the following, we will motivate and discuss potential applications of our method in active learning, creating pseudo-labels for training on unlabeled datasets and human-machine collaboration. In this section we used the majority vote of the uncalibrated ensemble as a prediction on the test sets as suggested in \cite{nnUNet}.

\begin{figure}
    \centering
    \includegraphics[scale=0.3]{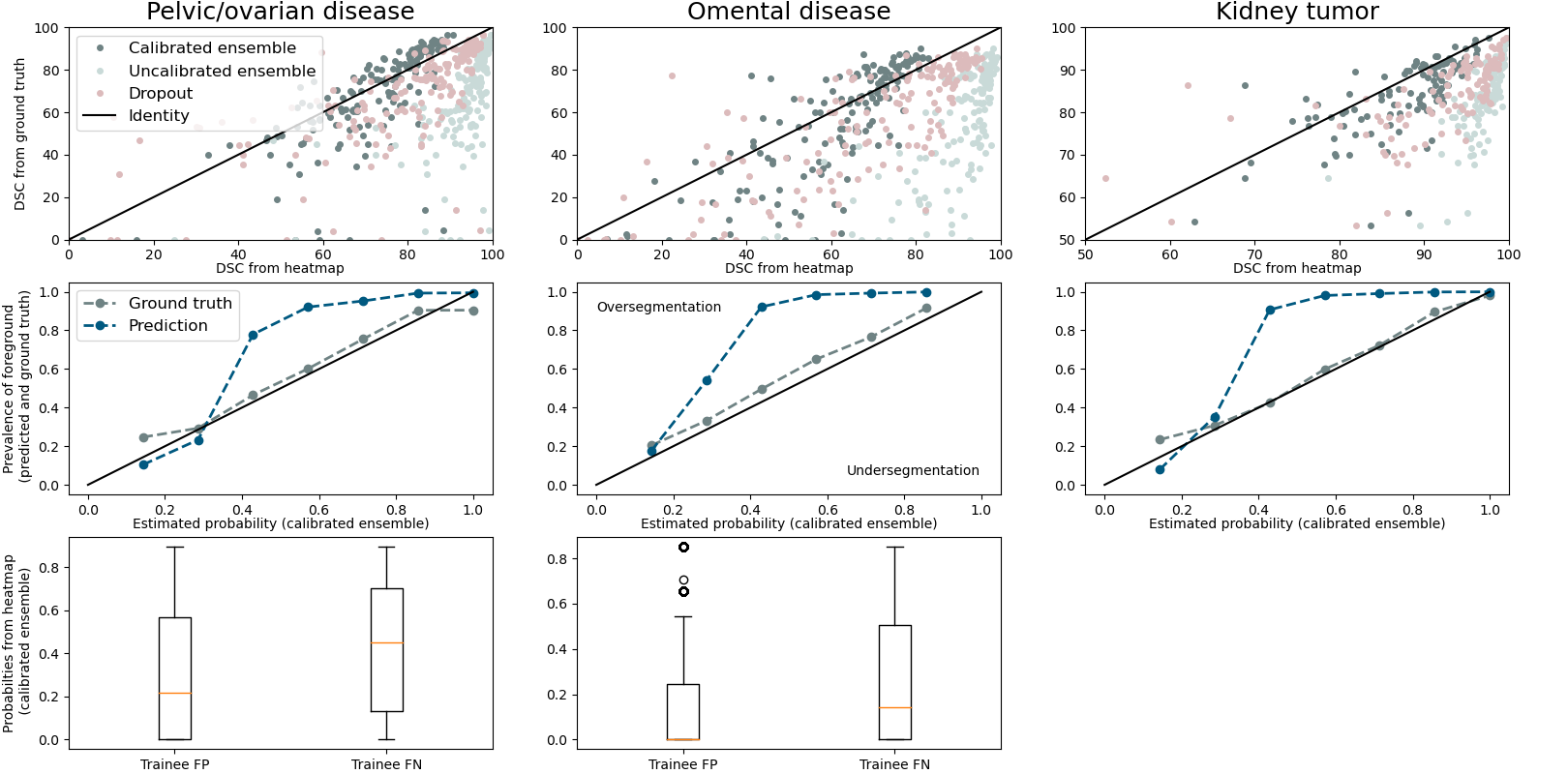}
    \caption{Motivational examples for usage in other areas of research. The first, second and third row motivate the usage in active learning, creating pseudo-labels on unlabeled images and human-machine collaboration respectively. The top row compares the estimated DSC, computed from different heatmaps, with the DSC computed from ground truth segmentations. Our approach demonstrates a better agreement between the two compared to uncalibrated ensembles and dropout-based methods. The middle row compares the estimated probabilities with the relative frequency of ground truth and predicted foreground. The large deviation of the two curves demonstrates the suboptimality of the prediction. A loss function that promotes similarity of the estimated probability and the relative frequency of predicted foreground might be applied on unlabeled datasets to solve this issue. The predictions were chosen to be a majority vote of predictions from the uncalibrated ensemble as these resulted in the higher DSCs compared to using calibrated ensembles or Dropout predictions for these tasks. The bottom row considers human-machine collaboration by checking the probabilities our approach computed in voxels where a trainee radiologist segmentation deviated from the ground truth.}
    \label{fig:applications}
\end{figure}

\subsection{Active learning}
Applying active learning to the process of manually labelling new datasets can bring huge advantages. It is either possible to directly reduce the manual labelling time by integrating a continuously updated model in the labelling process \cite{total_segmentator}, or to select a subset of scans that contribute most to the models performance \cite{AL_survey}.

To motivate the adoption of our suggested approach in the area of active learning we aimed at estimating the DSC using only a heatmap $h$ and prevalent binary predictions $p$. First, we compute the DSC of some prediction $p$ and the ground truth segmentation $s$ on the test set using the standard formulations of the DSC:

\begin{align*}
    TP(s,p) &= \sum_i s_i p_i \\
    FP(s,p) &= \sum_i (1- s_i)p_i \\
    FN(s,p) &= \sum_i s_i (1-p_i) \\
    DSC(s, p) &= 100 \times \frac{2TP(s,p)}{2TP(s,p)+FP(s,p)+FN(s,p)}.
\end{align*}

Following the hypothesis that the heatmaps created by our suggested approach do approximate the classification probability, we can compute the expected number of true positives, false positives and false negatives by simply replacing $s$ with $h$ in the formulas above. Hence, the DSC can be estimated with this technique without having access to the ground truth. The top row of Figure \ref{fig:applications} compares DSCs from the different heatmaps with the DSCs computed from the ground truth. %true DSCs.
It can be observed that the heatmaps created by uncalibrated ensembles and inference Dropout tend to overestimate the true DSC, while using the heatmaps created by our approach results in a more accurate DSC. The overestimation of the DSC is matches the observation from Figure \ref{fig:compare_heatmap} that suggested that uncalibrated ensembles and Dropout models tend to produce false positives and negatives with high confidence. %This suggests that the previously established approaches tend to underestimate the uncertainty while our approach is better suited for this task.

%Next, we investigated the usage of our method in the \textbf{creation of pseudo-labels for training on unlabeled datasets}. Let's assume heatmaps were computed on a large unlabeled dataset. Providing that the estimated probabilities are a good approximation to the prevalence of foreground in these scans, we can set a loss function that forces the output of a deep neural network to contain $p$ percent foreground across all regions where the heatmap estimated the classification probability as $p$. To check whether current predictions fulfil this, we computed the prevalence of ground truth and predicted foreground on the test sets using again Prazen Windowing with a linear kernel. The result can be found in the bottom row of Figure \ref{fig:applications}. The clear deviation of the two quantities suggests that the predictions could be improved by the loss function motivated above. The current predictions tend to annotate too much foreground in regions of large certainty. Additionally for the pelvic/ovarian lesions and kidney tumors it can be observed that the model tends to annotate too little foreground in regions of low certainty.

%write new description of the pseudo-label application
\subsection{Creation of pseudo-labels for learning on unlabeled datasets}
Next, we investigated the usage of our method in the creation of pseudo-labels for training on unlabeled datasets. Including unlabeled images in the training process can be of significant advantage compared to standard supervised methods, for example when using self-supervised methods as pretraining \cite{medical_self_supervised, self_supervised_wild} or using pseudo-labels in semi-supervised learning \cite{pseudo_labels, pseudo_labels2}.

As demonstrated in Figure \ref{fig:p_vs_p}, comparing prevalence of foreground voxels and predicted probability can be used to validate heatmaps generated by uncertainty quantification approaches. However, given a heatmap where prevalence of foreground voxels and predicted probability are reasonably well calibrated, this comparison can also be used to validate deep neural network predictions. Figure \ref{fig:p_vs_p} shows that only the calibration curve of the calibrated ensemble was relatively close to perfect calibrated on the test set, which is why we focused solely on this approach for this application. If the prevalence of \textit{predicted} foreground voxels does not match the probabilities generated by the heatmap, the prediction cannot be perfect. The second row of Figure \ref{fig:applications} demonstrates that this is not the case for our predictions which were trained in a fully supervised fashion. The current predictions tend to annotate too much foreground in regions of large certainty. Additionally for the pelvic/ovarian lesions and kidney tumors it can be observed that the model tends to annotate too little foreground in regions of low certainty.

It should be noted that even though a prediction method produces segmentations where the prevalence of foreground voxels and the predicted probability of a heatmap match, it must match with the ground truth. For example, a segmentation method that produces a checkerboard pattern in voxels with a probability of 0.5 would fulfill the optimality condition while producing biologically implausible segmentations. This is why methods cannot be trained only basing on this pseudo-label approach. Instead, we believe that it can either augment existing supervised approaches in a semi-supervised framework, or serve as a pre-training objective followed by supervised fine-tuning.

% Section on human-machine interaction
\subsection{Human-machine collaboration}
Lastly, we want to motivate the potential usage in human-machine collaboration. In this dicispline it is researched how and if humans can be partnered with machine learning systems to improve overall performance \cite{human_ai, human_AI_collaboration}

For this we used the unreviewed trainee radiologist segmentations that were available on a subset of 104 scans of the test set of the ovarian cancer data. More specifically, we wanted to understand if the AI can be used to potentially flag false positives or false negatives in the trainee's work. The last row of Figure \ref{fig:applications} shows boxplots of the estimated probabilities, generated by the calibrated ensemble, in voxels that were falsely annotated by the trainee radiologist as positive/negative. The plots show that the heatmap actually annotated a large number of false positive voxels with a very low probability. In fact, 50\% and 75\% of the false positive annotations of the trainee were annotated with a probability of less than 0.25 by the heatmap for pelvic/ovarian and omental disease respectively. Additionally, more than 25\% of the false negative annotations of the trainee were marked with a probability of greater than 0.7 by the heatmap. For the omental disease, more than 75\% of the false negatives were annotated with a probability of less than 0.5, indicating that the majority these mistakes by the trainee were also difficult to detect for the deep learning system.

It should be noted that simply removing trainee annotations with low probabilities can be harmful as also true positives of the trainee might be removed. Instead, such system could be used to flag potential false positives and negatives to the trainee and suggest to rethink the decision made.

\section{Discussion}

In this paper, we presented a simple, yet effective, approach for calibrating ensembles of deep learning models to approximate the classification probability in 3D medical image segmentation, while many other approaches do not scale well in these high-dimensional problems. We motivated why ensembles of identical models trained with different random seeds fail to approximate the classification probability and why models with differing sensitivity and precision are needed to do so. We demonstrated that it is feasible to apply the approach in complex segmentation problems using large 3D networks and that the resulting heatmaps approximate the classification probability. We also showed that the traditional approaches tend to produce false positives and negatives with high confidence. Considering the intersection/union of all segmentation produced by the different approaches, ours had the highest precision/sensitivity in all except one case. %To further improve these measurements on can simply increase the range of loss weights $w$ used in the calibrated ensemble. Future approaches might take this consideration into account when construction a set of loss weights $w$.
We further motivated how our method can be used in active learning \cite{AL_survey, total_segmentator}, computing pseudo-labels to learn from unlabeled data \cite{pseudo_labels, pseudo_labels2, medical_self_supervised, self_supervised_wild} and human-machine collaboration \cite{human_ai, human_AI_collaboration}.

Our main limitations are the following. While our method only considers the estimation of the classification probability, it does not contain any out-of-distribution detection \cite{ood}. However, we believe that such approaches can be combined with ours due to the simplicity of our suggested framework. Next, our currently suggested framework introduces high computational cost during inference making it unattractive for clinical deployment.

For future work we suggest investigating the combination of our approach with other UQ approaches such as out of distribution detection \cite{ood}. For this, we believe that it is possible to generalize our approach to consider also soft segmentations by adapting Equation \ref{eq:linear_system} adequately. To further improve the sensitivity and precision of the intersection/union of segmentations in the calibrated ensemble, it might be useful to adapt the choices of loss weights $w$ accordingly. We also believe that knowledge distillation might be of great help to reduce inference cost and facilitate clinical deployment. For this we believe that theoretical considerations and clinical implementations of our approach are needed to determine the value of UQ based methods like ours in human-machine collaboration. Furthermore, it would be interesting to see how our suggested method of pretraining on unlabeled datasets using pseudo-labels performs when being compared to other pretraining approaches like self-supervised learning \cite{medical_self_supervised, self_supervised_wild}. We also believe that our method could be extended to incorporate the impact of the uncertainty in the segmentation into Radiomics-based classification models using such segmentations, or even be extended from segmentation problems to classification problems like Radiomics-based classification.
%can be applied outside of segmentation, for example in radiomics-based medical classification models.

To conclude, we proposed a simple and intuitive approach to develop calibrated ensembles that approximate classification probabilities in 3D medical segmentation problems.

\section*{Acknowledgement}

The authors would like to thank everyone involved in the data collection and labeling process of the analyzed ovarian cancer datasets. In particular, the authors thank Lucian Beer, Vlad Bura, Hilal Sahin, Roxana Pintican and Marta Zerunian for assisting the manual labeling process. The authors also thank Naveena Singh, Anju Sahdev for collecting the patient data at the Barts Hospital, London, UK. Additionally, the authors thank Kathleen Barcy, Larry G. Maxwell, Nicholas W. Bateman, Thomas P. Conrads and John B. Freymann for the collection of the Apollo dataset, as well as Iris Allajbeu for the assistance in collecting the scans available from the Cancer Imaging Archive.

This work was partially supported by the Wellcome Trust Innovator Award [RG98755], The Mark Foundation for Cancer Research and Cancer Research UK Cambridge Centre [C9685/A25177], and the CRUK National Cancer Imaging Translational Accelerator (NCITA) [C42780/A27066]. Additional support was also provided by the National Institute of Health Research (NIHR) Cambridge Biomedical Research Centre (BRC-1215-20014). RW was supported by the Austrian Science Fund (FWF) [J4025-B26]. MC-O was funded by the EPSRC Tier-2 capital grant EP/P020259/1. The views expressed are those of the authors and not necessarily those of the NHS, the NIHR, or the Department of Health and Social Care.

The work by Öktem was supported by the Swedish Foundation of Strategic Research under Grants AM13-0049.

Microsoft Radiomics was provided to the Addenbrooke’s Hospital (Cambridge University Hospitals NHS Foundation Trust, Cambridge, UK) by the Microsoft InnerEye project.

\bibliographystyle{unsrt}
\bibliography{references}

\end{document}